\documentclass[lettersize,journal]{IEEEtran}
\usepackage{amsmath,amsfonts}
\usepackage{algorithmic}
\usepackage{algorithm}
\usepackage{array}
\usepackage[caption=false,font=normalsize,labelfont=sf,textfont=sf]{subfig}
\usepackage{textcomp}
\usepackage{stfloats}
\usepackage{url}
\usepackage{verbatim}
\usepackage{xcolor}
\usepackage{cite}
\usepackage{graphics}
\usepackage{graphicx}
\usepackage{comment}

\begin{document}

\title{A Novel Bioinspired Neuromorphic Vision-based Tactile Sensor for Fast Tactile Perception}

\author{Omar Faris$^{1,2}$,
        Mohammad I. Awad$^{1,3,4,*}$,
        Murana A. Awad$^{1}$,
        Yahya Zweiri$^{3,5,6}$,
        and Kinda Khalaf$^{1,3,4} $
\thanks{$^{1}$O. Faris, M.I. Awad, M.A. Awad, and K. Khalaf are with the Biomedical Engineering and Biotechnology Department, Khalifa University, P O Box 127788, Abu Dhabi, UAE  }
\thanks{$^{2}$O. Faris is with School of Computer Science, University of Lincoln, LN6 7TS, UK }
\thanks{$^{3}$M. I. Awad, Y. Zweiri, and K. Khalaf are with the Khalifa University Center of Autonomous Robotic Systems (KUCARS), Khalifa University, P O Box 127788, Abu Dhabi, UAE  }
\thanks{$^{4}$M. I. Awad, and K. Khalaf are with the Healthcare Engineering and Innovation Group (HEIG), Khalifa University, P O Box 127788, Abu Dhabi, UAE  }
\thanks{$^{5}$Y. Zweiri is with the Aerospace Engineering Department, Khalifa University, P O Box 127788, Abu Dhabi, UAE  }
\thanks{$^{6}$Y. Zweiri is with the Advanced Research and Innovation Center (ARIC), Khalifa University, P O Box 127788, Abu Dhabi, UAE  }
\thanks{$^{*}$Corresponding author: mohammad.awad@ku.ac.ae}
}

\markboth{Journal of \LaTeX\ Class Files,~Vol.~14, No.~8, August~2021}%
{Shell \MakeLowercase{\textit{et al.}}: A Sample Article Using IEEEtran.cls for IEEE Journals}

\IEEEpubid{0000--0000/00\$00.00~\copyright~2021 IEEE}

\maketitle

\begin{abstract}
Tactile sensing represents a crucial technique that can enhance the performance of robotic manipulators in various tasks. This work presents a novel bioinspired neuromorphic vision-based tactile sensor that uses an event-based camera to quickly capture and convey information about the interactions between robotic manipulators and their environment. The camera in the sensor observes the deformation of a flexible skin manufactured from a cheap and accessible 3D printed material, whereas a 3D printed rigid casing houses the components of the sensor together. The sensor is tested in a grasping stage classification task involving several objects using a data-driven learning-based approach. The results show that the proposed approach enables the sensor to detect pressing and slip incidents within a speed of 2 ms. The fast tactile perception properties of the proposed sensor makes it an ideal candidate for safe grasping of different objects in industries that involve high-speed pick-and-place operations.
\end{abstract}


\section{Introduction}

\IEEEPARstart{W}{hen} human beings pick objects, their fingers convey rich information about the grasping state through a set of tactile receptors that get activated by the physical touch between the hand and the object. With this sensory feedback, humans can easily execute complex grasping and manipulation operations in unstructured environments across objects with different properties. Equipping robotic systems with a similar capability of tactile perception may push the boundaries of various industries that involve grasping and handling a wide variety of objects and laborious manipulation tasks \cite{chen2018tactile, yousef2011tactile, li2020review}. Therefore, roboticists have put tremendous efforts towards developing artificial tactile sensors that can unlock the potential of deploying robots in such scenarios \cite{kappassov2015tactile}. 

Vision-based tactile sensors have seen increased popularity among researchers given their high dimensional and rich output that can reflect the physical properties of objects during the grasping process \cite{shimonomura2019tactile, yamaguchi2019recent}. These sensors employ a camera directed towards a skin that reacts to the interactions with the environment \cite{shah2021design, zhang2022hardware, li2023marker}. Notable vision-based tactile sensors that have been developed and extensively studied include GelSight \cite{yuan2017gelsight, abad2020visuotactile}, TacTip \cite{chorley2009development, lepora2021soft}, and FingerVision \cite{yamaguchi2019tactile}. The concepts behind these sensors have been further explored and improved to produce other variations that can excel in complex grasping and manipulation tasks such as contact shape detection \cite{donlon2018gelslim, zhang2022deltact}, shear force estimation during contact \cite{pang2021viko}, slip detection \cite{james2020slip}, light touch detection \cite{yamaguchi2021fingervision}, in-hand manipulation \cite{lambeta2020digit, she2021cable, wang2020swingbot}, and object localization \cite{ward2018tactip}. Despite their advantages and versatility, most of the existing vision-based sensors utilize frame-based cameras that have limited temporal resolution, rendering them inferior in applications where high speed operations are required \cite{muthusamy2020neuro}.

\IEEEpubidadjcol

\begin{figure}[t!]
\centering
\includegraphics[width=0.45\textwidth]{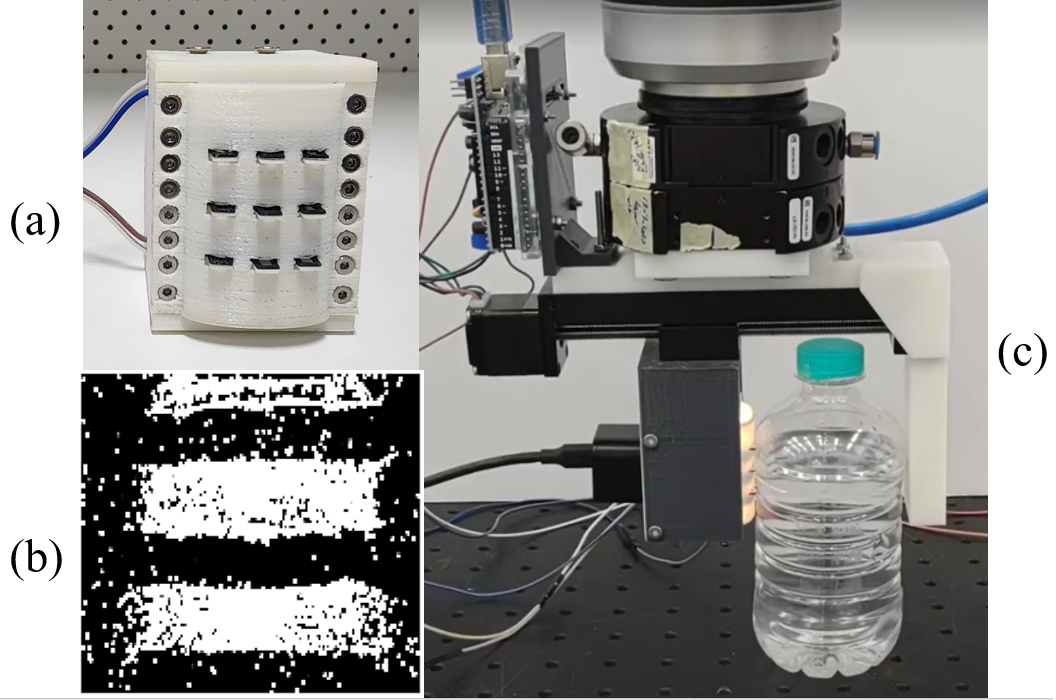}
\caption{The proposed Neuromorphic Vision-Based Tactile Sensor (a) the realized system, (b) the events (sensor's data) (c) the proposed sensor integrated on a robotic parallel gripper }
\label{introduction}
\end{figure}

Neuromorphic vision-based tactile sensors replace the conventional frame-based camera component of the sensor with an event-based camera. Event-based cameras take their inspiration from the human eye retina and generate asynchronous events, where each event is triggered at a pixel when the brightness intensity change exceeds a threshold, allowing them to have a high temporal resolution, low power consumption, and high dynamic range \cite{gallego2020event}. With these capabilities, event-based cameras have demonstrated superior performance compared to frame-based cameras in various applications that require high speed operations and low lighting conditions \cite{muthusamy2021neuromorphic}.  

Vision-based tactile sensors employing event-based cameras have been developed and utilized to achieve numerous complex sensing tasks. Notably, NeuroTac was developed based on the design principles of TacTip while replacing the frame-based camera with an event-based camera and was successfully employed in texture recognition \cite{ward2020neurotac} and edge orientation classification tasks \cite{macdonald2022neuromorphic}. Another tactile sensor employing an event-based camera proposed in \cite{kumagai2019event} was used to detect high speed phenomena, such as slip, during the contact between the sensor and the environment. A similar neuromorphic vision-based tactile sensor was also developed in \cite{sajwani2023tactigraph}. In the previous works, the sensor tip and markers are manufactured in separate processes, typically from elastomers, before joining them in a single body. Other works opted for using event-based cameras with silicone-based skins inside robotic grippers to detect incipient slip \cite{rigi2018novel} and measure forces and properties of objects during contact \cite{baghaei2020dynamic, naeini2019novel, huang2020neuro}.

More recently, Evetac tactile sensor \cite{funk2023evetac} was developed by combining an event-based camera with a variant of the skin seen in the GelSight sensor, which can be purchased off-the-shelf. The researchers demonstrated closed-loop controlled grasping of objects with fast detection of slip provided as a feedback to the controller.  Despite this variety, the development of existing neuormorphic vision-based tactile sensor is mostly approached by utilizing designs and skins that have been originally developed for frame-based cameras, which could be limiting the potential of such sensors.

\begin{figure*}[t!]
\centering
\includegraphics[width=0.85\textwidth]{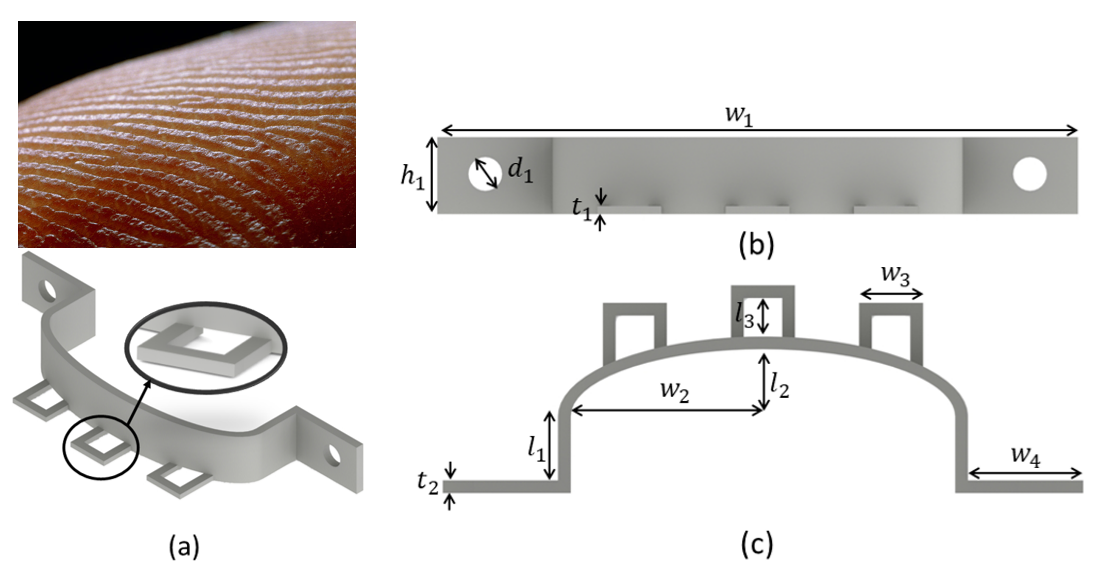}
\caption{ A 3D CAD render of a single bio-inspired marker in the finger: (a) Isometric view of the marker highlighting the marker protrusion. The protrusion design is inspired by the human finger ridges (b) Front view of the marker, and (c) Top view of the marker. The dimensions (in mm) are as follows: $h_1 = 4.80, d_1 = 2.10, w_1 = 40.00, w_2 = 12.00, w_3 = 4.00, w_4 = 7.20, t_1 = 0.50, t_2 = 0.80,  l_1 = 4.00, l_2 = 4.20, l_3 = 2.40$.}
\label{marker}
\end{figure*}

In this work, we propose a novel miniaturized bio-inspired neuromorphic vision-based sensor (illustrated in Fig. \ref{introduction}) that exploits the characteristics of event-based cameras to enable high-speed detection of physical phenomena during the grasping process. Inspired by the human fingerprint ridges, which have been linked with the sensitive tactile activity of the human hand \cite{jarocka2021human}, the sensor incorporates protruded markers that are triggered by light touch and minimal contact on top of a backbone body that deforms when the protruded markers reach it under heavy contact. The markers and the backbone body are manufactured using a flexible material that can be directly 3D printed with accessible and affordable commercial 3D printers. A small DVXplorer Mini neuromorhpic camera observes the deformation of the sensor skin under the lighting of two LED strips, whilst a rigid 3D printed case encapsulates all the components of the sensor together.

We demonstrate the capabilities of the sensor in a grasping stage classification task within a speed of 500 Hz (2 ms) using a data-driven approach that can learn the patterns of each grasping stage using training data from a single experimental run. The grasping stages include pressing, slip, and the absence of any activity between the sensor and the object. The approach is experimentally studied over a set of ten objects with varied properties. Overall, the contributions of this work can be summarized as follows:

\begin{itemize}
    \item Design and prototyping of a 3D printed bio-inspired neuromorphic vision-based tactile sensor.
    \item A data-driven approach for classifying the grasping stage using a single experimental run for any object.
    \item Experimental study of the grasping stage classification over a wide set of objects with different properties.
\end{itemize}

\section{Hardware Methods}

Our neuromorphic vision-based tactile sensor incorporates four separate sets of components: an event-based camera, a flexible skin with decoupled protruded markers, a rigid casing, and a pair of LED strip lights. The rigid casing attaches the components of the sensor together and can be connected to a desired robotic system.

\subsection{Event-based Camera}

Neuromorphic event-based cameras produce information in the form of spatio-temporal event spikes that occur as a response to the logarithmic brightness intensity changes in the scene. The events are generated asynchronously and independently at each pixel. A spatio-temporal event $e$ includes the spatial information of the pixel $x, y$, which reflects the location of the event, and the temporal information $t$, which represents the time the event occurred. 

Each event is generated when the absolute difference between the current and previous brightness intensity logarithm $L_I = log(I)$ at the corresponding pixel exceeds a threshold $C$:

\begin{equation}
\begin{split}
\bigtriangleup L_I(x,y,t) \: = \:
\mid L_I(x,y,t) - L_I (x,y,t- \bigtriangleup t) \mid
\\= Pol *\: C  \:\Bigg\vert \: C>0, Pol \in \{+1,-1 \}
 \label{eq1}
\end{split}
\end{equation}

where $\bigtriangleup t$ is the difference between the current time and the last time since an event was triggered at the corresponding pixel and $Pol$ is the event polarity. The polarity is determined by the change of the brightness at the pixel, where a positive polarity indicates an increase in the brightness, while a decrease in it generates a negative polarity.

\begin{figure}[t!]
\centering
\includegraphics[width=0.4\textwidth]{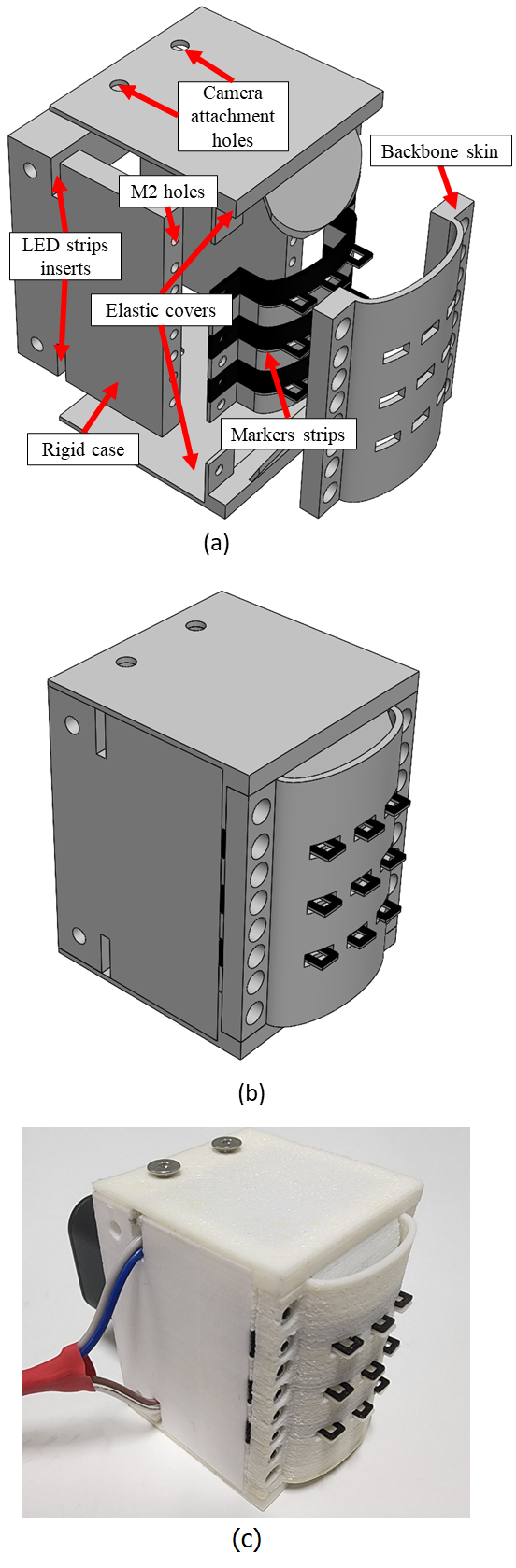}
\caption{Assembly of the proposed sensor (a) Exploded view of the sensor components and (b) Full assembly of the sensor body.  (c) physical model of the sensor}
\label{assembly}
\end{figure}

In this work, we utilize the state-of-the-art, commercially available DVXplorer Mini event-based camera as the main camera sensor. The DVXplorer Mini encompasses a high spatial resolution of 640 × 480 pixels in a compact body that is less than 30 mm in width and height. In this camera, the streams of events are produced at a temporal resolution of 5000 Hz (200 $\mu s$). An S-mount lens is attached to the camera to enable the adjustment of the camera focus for a better view of the sensor body. 

\subsection{Markers Design}

The bio-inspired markers in our proposed sensor are designed to extend beyond a flexible backbone skin. The markers are fabricated in the form of strips where each strip contains three protruded markers and a large single body that lies behind the backbone skin. The sensor houses six strips of markers, and each two strips are stacked together to form pairs. In total, the sensor contains nine pairs of protruded markers spread across six strips.

Given the use of an event-based camera, each pair of markers should have different colors in its design to cause a change in the brightness intensity when the markers move. Therefore, the strips of top markers are designed in a black color, whereas the strips of bottom markers are designed in a white color. The backbone skin is designed with a white color as it was experimentally found to be the color that minimizes noisy events in the absence of any tactile activity. The overall design and dimensions of the markers are shown in Fig. \ref{marker}. Each marker has a thickness of 0.5 mm and a width of 4.0 mm. The markers extend beyond the skin of the strip by around 3.20 mm ($t_1 + l_3$ in Fig. \ref{marker}c).  

When attached to the backbone skin, this protrusion is reduced to 2.00 mm given the fact that the backbone skin is designed with a thickness of 1.20 mm. The backbone skin allows the protrusion of markers through rectangular holes across its structure. The backbone skin and the strips of markers include holes that fit an M2 sized screw such that they are conveniently attached to the rigid casing.

With the proposed markers design, the sensor can convey information about light contact and fast phenomena during the interaction with objects through the protruded sections of the markers, mimicking the way finger ridges work. The stacked design of white and black color triggers events whenever the protrusions of the markers move. When the contact between the sensor and the object increases, the markers protrusions move towards the body of the sensor until they become on the same level of the backbone skin. Afterwards, the backbone skin becomes in direct contact with the object, allowing the sensor to continue conveying information about the scene interactions even under heavy contact. We believe that this multi-level sensing paradigm can be utilized to optimize the performance of the tactile sensor for 

\subsection{System Assembly}

The rigid casing attaches the remaining parts of the sensor together in a single body. The rigid casing includes a space to house the camera and LED strip lights. The elastic parts of the sensor, which include the backbone skin and the strips of markers, are mounted and screwed at the front side of the rigid casing. The casing also includes holes to screw the event-based camera as the body of the DVXplorer Mini camera already comes with mounting holes. The distance between the markers and the camera was determined experimentally before manufacturing the rigid casing to ensure that the camera fully observes the internal region of the markers and backbone skin.

Internal LED strip lights are glued at the top and bottom of the rigid casing to provide consistent illumination inside the sensor and ensure the repeatability of operation with the event-based camera. Finally, flexible white skins are designed to cover the exposed top and bottom parts of the sensor such that the effects of external lighting conditions on the camera are minimized. Fig. \ref{assembly} demonstrates 3D computer-aided models (CAD) of the components of the sensor and their assembly.

\subsection{Fabrication Process}

Fused deposition modelling (FDM) 3D printing technology was utilized to fabricate the sensor body. The markers were designed to be printed using the NinjaFlex Thermoplastic Polyurethane (TPU) material, whereas the rigid body is printed using Polylactic Acid (PLA) material. Off-the-shelf LED strip lights are glued inside the rigid casing, whereas M2 and M3 screws are used to attach the sensor components together, including the body parts and the DVXPlorer Mini camera.

Existing skin designs in neuromorphic vision-based tactile sensors generally rely on silicone elastomers and involve elaborate and multi-step fabrication processes to ensure that the markers and the skin of the sensor tip are made from different colors and appropriate sizes. The major alternative, which is found in the Evetac sensor \cite{funk2023evetac}, is to purchase commercially available skins of existing sensors. Such a choice can be expensive and limits the operability of the sensor to the presence of a material that may not be always accessible. Our adoption of the FDM technology and NinjaFlex material for the flexible parts of our sensor alleviate all these concerns and enable rapid, single-step, affordable, reliable, and simple prototyping of the sensor body.

\section{Software Methods}

\begin{figure}[b!]
\centering
\includegraphics[width=0.475\textwidth]{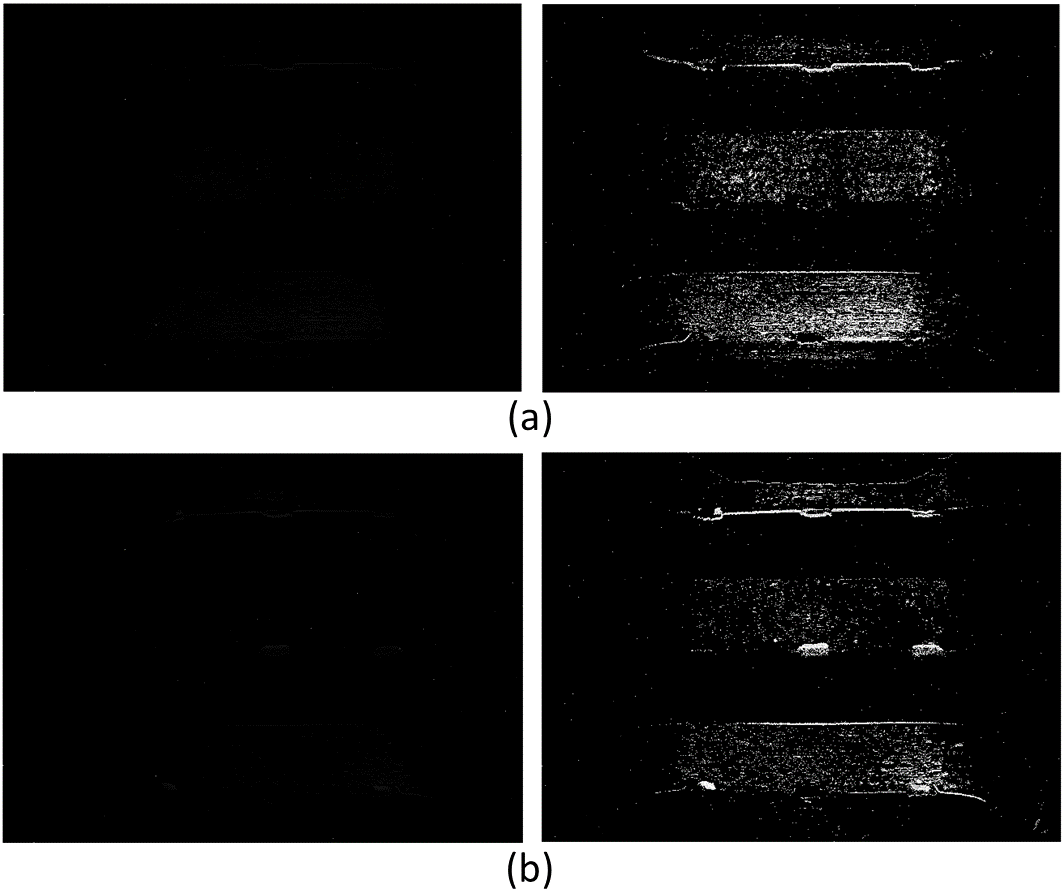}
\caption{Sample frames before (left) and after (right) applying the sigmoid function for noise reduction during (a) pressing and (b) slippage.}
\label{noise}
\end{figure}

When event-based cameras are used to achieve a task, a single stream of events may not provide enough information to reach the desired performance. Events are usually acquired and processed through specific representations or encoding methods to obtain a relevant representation \cite{gallego2020event}. Subsequently, a range of methods can be utilized to output the desired parameters from the constructed representations. In this work, we opt for processing the triggered events as a single frame to use it with existing learning-based data-driven convolutional neural networks (CNN).

\subsection{Events Pre-processing}

We group the streams of events happening over a sliding window of the past $h_t$ milliseconds with an update rate of 2 ms regardless of the events polarity. In other words, at each 2 ms, the group of events in the sliding window is updated by replacing the oldest 2 ms of events with the newest 2 ms of events. The events are then processed to form an image that encodes their spatio-temporal information through a heatmap-based representation that penalizes noisy pixels. In details, the final heatmap representation of the events occurring over a single sliding window is generated through the following steps:

\subsubsection{Events Aggregation}

Firstly, a sliding window is constructed over the past group of events. At a timestamp $t_c$, each pixel $x, y$ has a total number of events $E$ that equals the number of events that were triggered in that pixel over the past $t_h$ ms:

\begin{equation}
E_{x,y}(t_c) = \Sigma_{t=t_c-t_h}^{t=t_c} e_{x,y}(t)
\label{eq2}
\end{equation}

where $t_c$ is updated every 2 ms in time. We test $t_h$ over a range of values between 2 ms and 50 ms. In the case of $h_t = 2\ ms$, the instantaneous heatmap is obtained without any previous history of the events. Subsequently, a dummy frame is constructed to encode these events. where each pixel represents the number of events that happened in this pixel within the specified sliding window.

\subsubsection{Noise Reduction}

Event-based cameras are known for their generation of high amount of noisy events even in the absence of any activity or motion. Denoising methods and feature detectors, such as corner detectors, are often used to reduce the noise in the acquired events. However, these methods usually have poor performance when small timestamps are used. Alternatively, we observed that the noisy pixels in the DVXplorer Mini generate a significantly higher number of events than the useful pixels in a 2 ms timeframe. Therefore, we resort to the penalization of pixels with high number of events to reduce the noise. This is achieved by the pixel-wise multiplication of the dummy frame with the sigmoid function.
A sigmoid function maps an input $a$ by a parameter $b$ such that when the value of $a$ is equal to $b$, the output of the sigmoid function is 0.5. When the values of $a$ are larger than $b$, the sigmoid output converges to 1, whereas it converges to zero when the values of $a$ are lower than $b$:

\begin{equation}
    \sigma(a, b) = \frac{1}{1 + e^{-(a - b)}}
\label{eq4}
\end{equation}

Given these properties, we utilize the sigmoid function to penalize pixels that generate events that are much higher than the average number of events generated across the pixels. Therefore, we choose $a$ to be the negation of $E_{x,y}(t_c)$ (i.e., $a = -E_{x,y}(t_c)$) such that the values higher than $b$ approach zero instead of one, while the values lower than $b$ approach one instead of zero. For the value of the parameter $b$, we found by trial and error that choosing it to be the average number of events in triggered pixels across the dummy frame yielded the desired noise reduction effect. In other words, for each dummy frame $E_{x,y}(t_c)$, we sum the values of non-zero pixels and divide it by the number of non-zero pixels in the frame and assign that value to the parameter $b$.

\begin{figure*}[t!]
\centering
\includegraphics[width=0.95\textwidth]{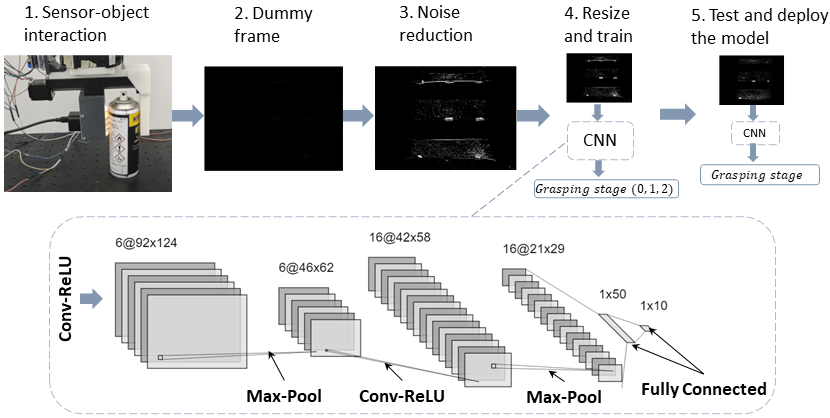}
\caption{Flow of the proposed approach for processing the events and classifying the heatmap-based frames, along with the structure of the CNN used for the grasping stage classificaitons.}
\label{cnn}
\end{figure*}

After calculating the pixel-wise value of the sigmoid function based on the aforementioned parameters, we multiply it pixel-wise by the original values of $E_{x,y}(t_c)$ such that pixels with high number of events are reduced to near zero:

\begin{equation}
D_{x,y}(t_c) = E_{x,y}(t_c) * \sigma_{x,y}(-E_{x,y}(t_c), b)
\label{eq3}
\end{equation}

Fig. \ref{noise} shows samples of frames with reduced noise based on the described method during pressing and slippage. Before applying the sigmoid function, a few noisy pixels produce a far higher number of triggered events, resulting in an image that is almost black when normalized. After the sigmoid function is applied, the values of the noisy pixels are reduced to near zero as they are much higher than the average number of non-zero pixels events across the image. Hence, the noise-reduced image highlights the useful information from the triggered events in the scene.

\subsubsection{Final Heatmap}

A remapped heatmap is constructed by remapping the values of all pixels such that they are between 0 and 255 through the pixel-wise division of the noise-reduced dummy frame by the highest value of any pixel and multiplying the resultant values by 255. Then, the remapped heatmap resolution is reduced by a fifth such that the final heatmap has a resolution of 128 × 96 pixels instead of the original camera resolution (i.e., 640 × 480 pixels). These two operations can be summarized as follows:

\begin{equation}
H(t_c) = resize(map(D(t_c), 0, 255), 0.2)
\label{eq5}
\end{equation}

\subsection{Deep Learning Architecture}

After constructing the final heatmap, a convolutional neural network (CNN) to utilized to achieve the desired task. The proposed network follows the simple LeNet architecture \cite{lecun1998gradient}. The input to the network is the constructed heatmap for each timestamp, whereas the output determines whether the heatmap reflects the absence of any activity, pressing of an object, or an ongoing slip. Fig. \ref{cnn} highlights the pipeline of the heatmap generation and the detailed structure of the proposed CNN.

\section{Experiments}

We examine the performance of the sensor in a challenging task of early classification of grasping stages which include the absence of any activity, object pressing, and object slippage. Early detection of robot-object interactions are essential for high-speed operations and accurate control crucial for successful grasping. In particular, slip represents a significant challenge in pick-and-place operations as it has to be detected as soon as possible to prevent the object from falling.

\begin{figure}[t!]
\centering
\includegraphics[width=0.48\textwidth]{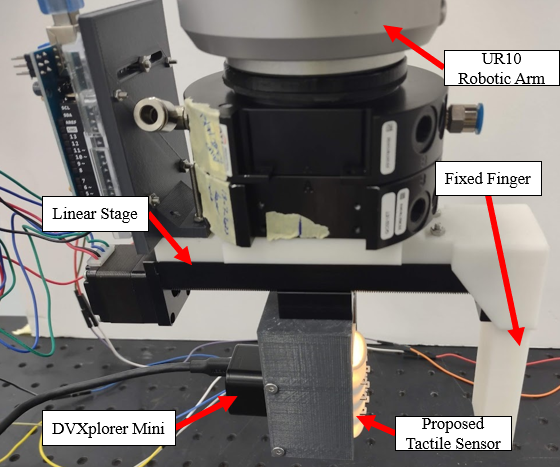}
\caption{Experimental setup and its components used for collecting the data and testing the proposed sensor and the classification method}
\label{exp_setup}
\end{figure}

To perform the experiments, we attach the sensor to a miniature moving stage with an extra fixed finger to create a parallel grasping setting. The gripper is subsequently attached to the UR10 to proceed with the experiments. We use Robot Operating System (ROS) middleware to communicate with all the involved components in the experimental setup. Fig. \ref{exp_setup} illustrates the experimental setup used to perform the experiments.

\begin{figure}[t!]
\centering
\includegraphics[width=0.48\textwidth]{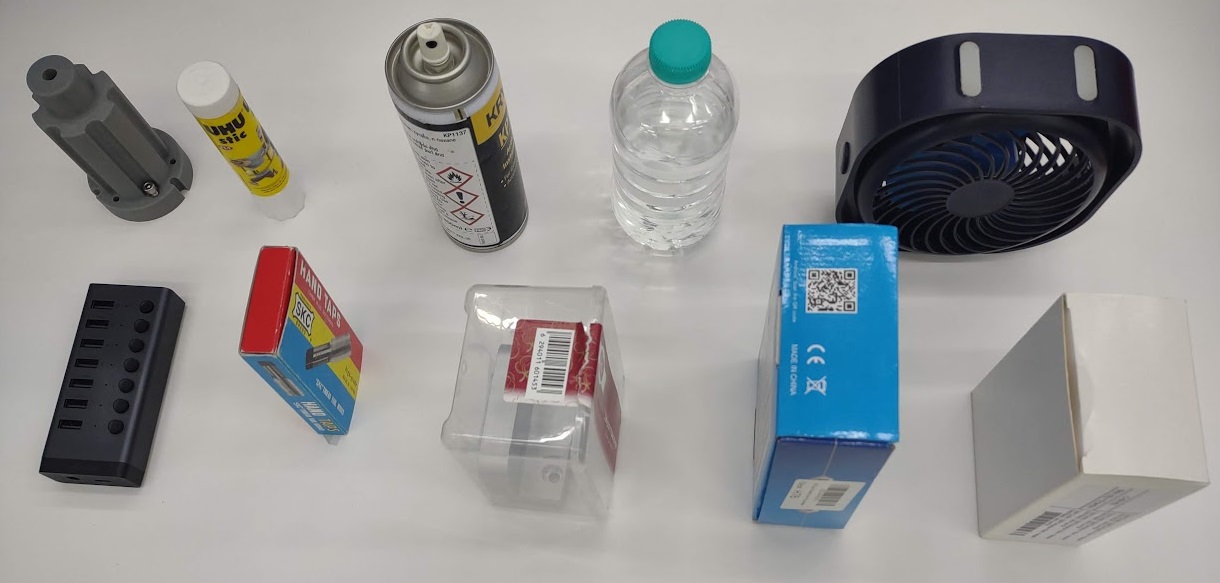}
\caption{The ten objects that are used to perform the experiments. From the top left corner in a clockwise direction: a 3D Printed Object with irregular surface, an UHU glue stick, a sprayer bottle, a water bottle, a fan, a white box, a textured box, a plastic container, a small box, and a USB hub.}
\label{objects}
\end{figure}

The experiments involve testing the developed algorithm for grasping stage classification over a wide range of objects. The objects were chosen to have contrasting properties including different levels of translucency, reflectivity, weight, sizes, and surface shape and irregularity. The final set of objects is shown in Fig. \ref{objects}.

Each experimental run on each object starts with a light press on the object being tested, followed by a command for the robot to lift the object. During the pressing, we ensure that the applied force for each object is not enough to grasp it. Therefore, the gripper will fail to lift the object and slip will be induced between the sensor and the object. We repeat the same experimental protocol ten times for each object. We use the heatmap-based representation and CNN structure described in the previous section to generate the data required to train and test the classification of the grasping stages during each experiment.

To generate ground truth labels for the collected data, we utilize observations from the event-based camera, the UR10 robot, and recorded videos of all experiments. For the pressing, we match the time at the start of the pressing process with the generated events from the DVXplorer Mini to mark the start of the interaction between the object and the sensor. In a similar manner, we use the timestamps from the lifting motion of the UR10 robotic arm and match it with the moment the event-based camera starts to generate events from the object slippage. The absence of any activity are randomly sampled from the remaining timestamps. To balance the dataset, the labels from all classes are reduced by random sampling to match the number of labels in the least present class. If the pressing or slippage continues longer than 500 ms, only the first 500 ms are considered to reduce the amount of data and focus on the most relevant and useful information.

\section{Results}

We study and analyze the collected data in three different ways. Firstly, we start by training the proposed CNN structure on a single experimental run from each object while keeping the remaining nine experimental runs to test whether the network can learn pressing, slippage, and absence of activity patterns for the objects with data from a single experimental run. From the trained network, we study the effect of varying the events history parameter $t_h$ from eq. \ref{eq2}. We test a range of values that include the instantaneous heatmap (i.e., $t_h = t_c = 2\ ms$) and the history of the past 10 to 50 ms of events with a step of 10 ms. We compare the average accuracy of classification across the ten objects for each history of events to determine the value of events history that shows the highest accuracy.

After finding the best value for the events history parameter, we extend the analysis on the performance of the network on each object by reporting by reporting the average time required to achieve a correct classification for each class (i.e., whether the network can detect the first moment of slippage and pressing) and having a look at the predictions of one full experimental run. Finally, we test the generalization of the network to unseen objects by limiting the training data to a single experimental run from five objects, while keeping all the data from the remaining five objects to test the generalization of the network and its ability to adapt to new objects with different properties. We compare the results of this test against what the network predicts when it is trained on all objects.

\subsection{Effect of Longer Events History}

\begin{figure}[t!]
\centering
\includegraphics[width=0.48\textwidth]{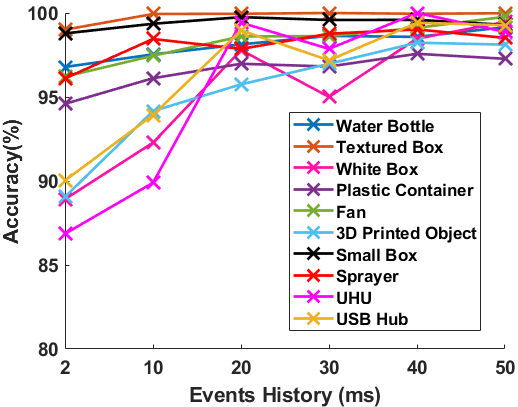}
\caption{Average accuracy for the nine test experimental runs for each object across six different windows for the history of events.}
\label{hist}
\end{figure}

Fig. \ref{hist} demonstrates the effect of increasing the history of events taken into account when training the model. Each point demonstrates the average classification accuracy for the nine experimental runs left for testing the model for each object. Instantaneous heatmaps with $t_h = 2\ ms$ results in the worst performance across all objects. Increasing the history of events improves the performance albeit on a varying level depending on the object, with the improvements becoming marginal after 20 ms of events history. We choose $t_h$ to be 40 ms to carry out the remaining analysis as it is the value of events history that provided the best performance for five objects out of the ten. On the other hand, three objects had their best average accuracy when $t_h = 50\ ms$, while two objects showed the highest average accuracy for a $t_h = 20\ ms$.

\begin{figure}[t!]
\centering
\includegraphics[width=0.425\textwidth]{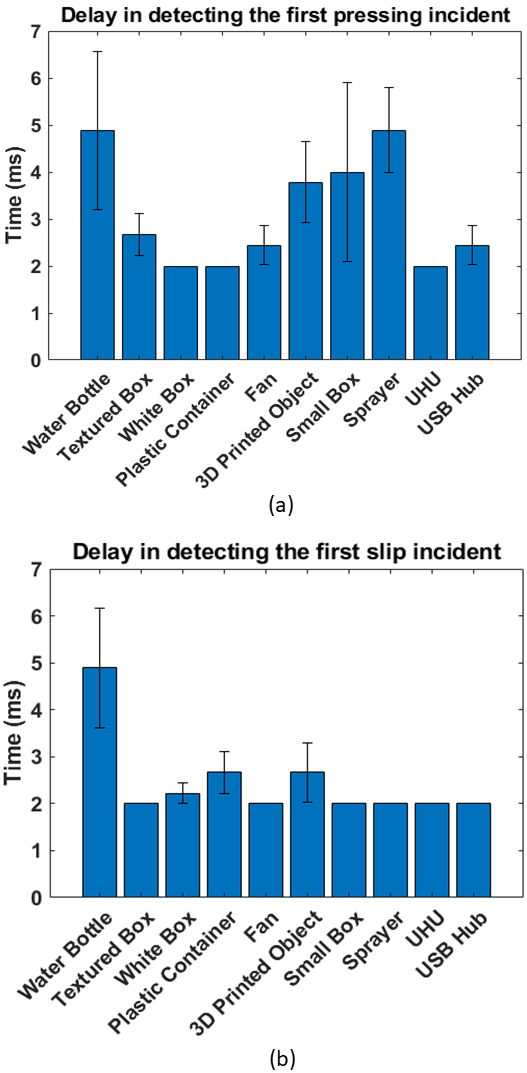}
\caption{Average delay of the nine test experimental runs of each object and the corresponding error bars representing the standard mean error in detecting the first incidence of (a) pressing and (b) slippage.}
\label{delay}
\end{figure}

\begin{figure}[t!]
\centering
\includegraphics[width=0.475\textwidth]{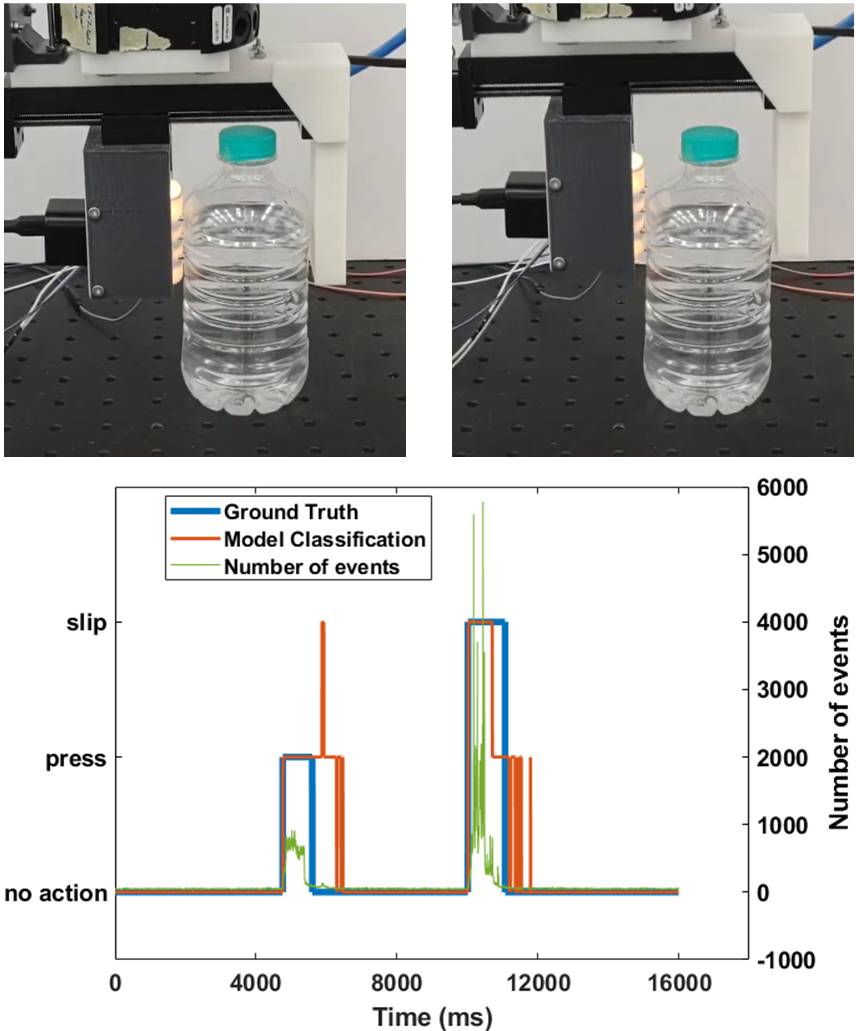}
\caption{A test experimental run for the Water Bottle object showing the grasping stage classification from the beginning to the end.}
\label{full_pred}
\end{figure}

\subsection{Learning from a single experimental run}

For $t_h = 40\ ms$, our method produces highly accurate grasping stage classifications across all objects in the test set when the network is trained on a single experimental run for each object. The average accuracy for the 9 test runs for all objects exceeds 95\%. This means that our proposed approach allows the proposed tactile sensor to detect a pressing action and the presence of a slip in any object once the data from a single experimental is provided to train the network. 

Fast detection of the interactions between the sensor and the objects ensures a safe grasping process. In particular, detecting slip is necessary to prevent the object from falling. Fig. \ref{delay} highlights the delay in detecting pressing and slip incidents, where a delay of 2 ms demonstrates a correct detection of the incident from the first moment. The figure shows the average delay across the nine test experimental run and the corresponding error bars that represent the standard error. Our approach succeeds in providing an average delay of less than 5 ms for both pressing and slippage detection for all objects. When detecting slippage, the sensor detects the slip with minimal delay across all trials in six out of ten objects.

Fig. \ref{full_pred} demonstrates the full predictions of the grasping stage by our trained model for one test experimental run for the Water Bottle. During the beginning of pressing and slippage, the model provides highly accurate classifications. However, near the end of each incident, the model performance deviates from the ground truth. This could be attributed to several factors. Firstly, the ground truth is manually labelled based on the number of detected events, which does not include the after effects of the pressing or slippage incidents that may still cause motion in the markers. Secondly, the training set from most objects only included the first 500 ms of each incident, whereas the shown predictions demonstrate predictions across the whole experimental run containing information the model has not been trained on. Lastly, the interactions between the object and the sensor are more complex than a single classification. Object motion similar to slippage may occur during the pressing, while a motion of the markers similar to the pressing may also occur during the object slippage depending on the motion of the object inside the gripper. This may cause the model to confuse the grasping stage and produce wrong predictions.

\subsection{Generalization to unseen objects}

\begin{figure}[t!]
\centering
\includegraphics[width=0.475\textwidth]{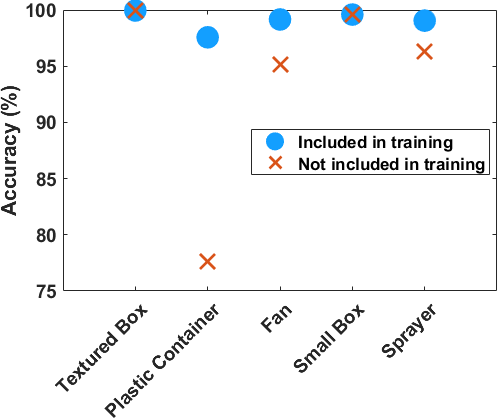}
\caption{Average accuracy of the grasping stage classifications when the data from five objects are not included in the training set, along with the corresponding performance when the data are included for a single experimental run.}
\label{unseen}
\end{figure}

Five objects were chosen randomly to be removed from the training dataset. These objects are the textured box, the plastic container, the fan, the small box, and the sprayer. Fig. \ref{unseen} compares the average classification accuracy for the five objects when the network is not trained on their data against the average accuracy when they are included in the training set. 

As expected, the model generally provides a better performance when the data from an object is included in the training set. Nonetheless, for most unseen objects, the model delivers highly accurate grasping stage classification that is comparable to the performance when the objects are included in the training set. The only exception in the studied objects is the plastic container, where the accuracy was significantly lower, potentially due to the fact that the plastic container possesses considerable differences in properties relative to the other objects involved in the training. The surface shape of the plastic container touching the sensor causes  irregular deformations in the sensor skin that are not present when other objects touch the sensor.

\section{Conclusion}

In this work, we presented a novel bio-inspired vision-based sensor that utilizes a neuromorphic event-based camera to observe the deformation of a 3D printed skin for fast inference of information during tactile interactions with the environment. The 3D printed skin is inspired by the fingerprint ridges and designed with a stacked set of protruded markers with different colors to trigger events whenever the markers move, allowing for fast tactile perception caused by the light touch of the sensor skin. We experimentally studied the proposed sensor on a grasping stage classification task, where the sensor was tested on its ability to classify during a grasping task whether the sensor is pressing on the object, the object is slipping, or the absence of any activity. We developed a heatmap-based representation to encode the event-based data from the camera and classify it using a CNN-based neural network. Experimental testing of the sensor in the grasping stage classification task with ten objects revealed that our sensor can perform highly accurate classifications of pressing and slippage actions within a speed of 2 ms with data when trained on a single experimental run per object, while also delivering promising results for generalizations towards unseen objects.

In the future, we plan to utilize the sensor for developing fast detection of objects properties such as object stiffness and texture. Additionally, we plan to explore new approaches that can improve the speed of slip detection and detect incipient slip. This will enable our sensor to be deployed in high-speed pick-and-place and object sorting applications.

\section*{Acknowledgement}


\end{document}